# A Deep Learning Approach to Language-independent Gender Prediction on Twitter

**Reyhaneh Hashempour**
University of Essex
rh18456@essex.ac.uk

**Barbara Plank**
IT University of Copenhagen
bplank@itu.dk

**Aline Villavicencio**
University of Essex
avill@essex.ac.uk

**Renato Amorim**
University of Essex
r.amorim@essex.ac.uk

## Abstract

This work presents a set of experiments conducted to predict the gender of Twitter users based on language-independent features extracted from the text of the users' tweets. The experiments were performed on a version of TwiSty dataset including tweets written by the users of six different languages: Portuguese, French, Dutch, English, German, and Italian. Logistic regression (LR), and feed-forward neural networks (FFNN) with back-propagation were used to build models in two different settings: Inter-Lingual (IL) and Cross-Lingual (CL). In the IL setting, the training and testing were performed on the same language whereas in the CL, Italian and German datasets were set aside and only used as test sets and the rest were combined to compose training and development sets. In the IL, the highest accuracy score belongs to LR whereas in the CL, FFNN with three hidden layers yields the highest score. The results show that neural network based models underperform traditional models when the size of the training set is small; however, they beat traditional models by a non-trivial margin, when they are fed with large enough data. Finally, the feature analysis confirms that men and women have different writing styles independent of their language.

## 1 Introduction

Gender prediction is a sub-task of author profiling where given a text written by an author, the goal is to predict the author's gender. Author profiling can benefit different domains. For example, the marketing department of a company can use it to learn about the gender or age of anonymous reviewers who have written on product review websites and react accordingly. Another example could be in forensic linguistics where people hide their real identity (e.g. gender) on the internet and we want to reveal it.

Most existing work has focused on linguistic features of texts (Schler et al., 2006; Cheng et al., 2011; Schwartz et al., 2013) and few included non-linguistic features such as metadata (Plank and Hovy, 2015; Ljubešić et al., 2017). The goal of the project is twofold: 1) to build a gender prediction model which is independent of the language a text is written in and 2) to compare the performance of traditional machine learning with deep learning. The former helps with the progress of social media analysis without relying on a specific language and the latter contributes to a better understanding of potentials and limitations of deep learning.

The dataset is a version of TwiSty corpus(Verhoeven et al., 2016) that contains gender annotations for a total of 6,482 authors posting in Portuguese, French, Dutch, English, German, or Italian. Portuguese has the biggest and Italian the smallest number of users with 3,066 and 306 users respectively. There is an equal number of male and female users for each language. Each user is represented by 200 tweets labelled with the gender.

## 2 Features and Experimental Setup

Three groups of features were selected including: Structural e.g. ***Average Length of tweets***; Punctuation: *The number of punctuation marks per sentence*, e.g. **?** *(question mark)* and Special Characters e.g. **#** *(hashtag)*. Structural features were defined based on the idea that women tend to write longer (Cheng et al., 2011). Punctuation was also deemed important since according to Mulac et al. (1988) women tend to use more question marks. Moreover, some punctuation marks, e.g. **: *(colon)*** or **; *(semicolon)*** are parts of emoticons which are also considered good gender predictive features (Ljubešić et al., 2017).

| Lang | Ins | LR | FFNN1 | FFNN2 | FFNN3 |
|---|---|---|---|---|---|
| PT | 3066 | **64.24** | 55.70 | 56.51 | 54.40 |
| FR | 1008 | **70.30** | 51.98 | 52.97 | 50.00 |
| NL | 894 | **67.29** | 55.87 | 54.19 | 43.02 |
| EN | 850 | **64.71** | 46.47 | 47.65 | 54.71 |
| DE | 358 | **67.59** | 48.61 | 63.89 | 59.72 |
| IT | 306 | **58.70** | 45.16 | 43.55 | 56.45 |

Table 1: Gender classification results on six languages in the Inter-language (IL) setting.

The models were trained on standardized (zero mean, unit variance) language independent features described above with logistic regression (LR) and three Feed Forward Neural Networks with back propagation (FFNN) with one, two, and three hidden layers where LR was used as the baseline. The experiment was conducted in two different settings: inter-language (IL) and cross-language (CL). In the IL setting, the training and testing were performed on the same language. In other words, if a model was trained on tweets in French, it was tested on tweets in French too; whereas in the CL setting, the model was trained on a dataset containing tweets in languages other than Italian and German and then tested on two datasets that included tweets in German or Italian respectively. There is an equal number of males and females in the datasets, therefore, accuracy was safely used as the performance metric.

## 3 Result and Conclusion

| Lang | Ins | LR | FFNN1 | FFNN2 | FFNN3 |
|---|---|---|---|---|---|
| DE | 358 | 57.26 | 77.09 | 79.89 | **83.52** |
| IT | 306 | 59.48 | 76.80 | 79.08 | **85.62** |

Table 2: Gender Classification results in Cross-language setting.

Table 1 demonstrates the results achieved for gender classification in the IL setting. An overall glance at the table suggests the superiority of LR to all FFNN models by a noticeable margin; however, such a conclusion cannot be drawn due to the small size of the datasets and the fact that neural networks require big data to learn an effective and robust model. This fact is more evident when we compare the performance of FFNN models trained on Portuguese with the ones trained on Italian considering the difference in size between these datasets. Portuguese dataset is large enough to learn an FFNN model which performs more consistently, with 0.86% standard deviation (less than 1%), while the results produced by FFNN models trained and tested on Italian are very variable showing 5.7% standard deviation.

Table 2 presents the accuracy scores of the models in the CL setting. FFNN with three hidden layers (FFNN3) shows the best performance with the accuracy scores of 85.62% and 83.52% for Italian and German respectively. The results are totally in line with what is expected from the neural network based models where a larger networks trained with more data is expected to perform better.

The results also show that tweets of each gender have special features that differentiate them from those of the other gender. This confirms other researchers' work (Schler et al., 2006; Cheng et al., 2011; Schwartz et al., 2013; Ljubešić et al., 2017). Furthermore, since in the CL setting, the model was trained and tested on totally different language sets, the features can be considered language-independent.

Despite the promising results in this work, some areas still need to be explored. For example, the annotation of the dataset was based on users' profile pictures and self-reports. We know that some people tend to hide their identity on the web and might use fake pictures. Therefore, the results obtained from the models tested on on-line datasets should be interpreted with caution. One way to tackle this problem could be testing the models on texts whose author's gender is certainly known.